\documentclass[conference]{IEEEtran}
\IEEEoverridecommandlockouts
\usepackage{cite}
\usepackage{amsmath,amssymb,amsfonts}
\usepackage{algorithmic}
\usepackage{graphicx}
\usepackage{textcomp}
\usepackage{xcolor}
\usepackage{url}
\usepackage{booktabs}

\def\BibTeX{{\rm B\kern-.05em{\sc i\kern-.025em b}\kern-.08em
    T\kern-.1667em\lower.7ex\hbox{E}\kern-.125emX}}
\begin{document}

\title{FlightSense: An End-to-End MLOps Platform for Real-Time Flight Delay Prediction via Rotation-Chain Propagation Features and Agentic Conversational AI\\

}

\author{\IEEEauthorblockN{1\textsuperscript{st} Aditi J. Shelke }
\IEEEauthorblockA{\textit{Stevens Institute of Technology} \\
Hoboken, NJ, USA \\
aditishelke333@gmail.com}
\and
\IEEEauthorblockN{2\textsuperscript{nd} Renuka J. Shelke}
\IEEEauthorblockA{\textit{Axtria Inc.} \\
Berkeley Heights, NJ, USA \\
renukashelke1999@gmail.com}
\and
\IEEEauthorblockN{3\textsuperscript{rd} Yash M. Kamerkar}
\IEEEauthorblockA{\textit{Stevens Institute of Technology} \\
Hoboken, NJ, USA\\
ykamerka@stevens.edu}
\and
\IEEEauthorblockN{4\textsuperscript{th} Nitin Hazarani}
\IEEEauthorblockA{\textit{Meta} \\
Menlo Park, CA, USA \\
hazaraninitin@gmail.com}
}

\maketitle

\begin{abstract}
Flight delays impose cascading operational and financial burdens across the aviation network, costing the U.S. economy billions of dollars annually by disrupting interconnected aircraft rotation systems. While prior machine learning approaches have demonstrated strong predictive performance, most treat upstream delays as static input variables rather than explicitly modeling how delays propagate dynamically through aircraft rotation chains, and none have deployed such systems alongside a live weather-aware conversational AI interface for end-user interaction.
 
This paper presents FlightSense, an end-to-end MLOps platform for real-time flight delay prediction built through a progressive three-version feature engineering framework. Version~1 trains an XGBoost classifier on 11 schedule-based features --- airline, origin-destination pair, distance, air time, and holiday indicators --- establishing a baseline ROC AUC of 0.732 on 7.07 million BTS 2018 On-Time Performance records. Version~2 introduces 11 delay propagation features derived from aircraft rotation chains via tail-number tracking --- including previous flight arrival delay, turnaround time, tight turnaround indicator, and cumulative daily aircraft delay --- yielding the dominant performance gain (AUC 0.732 $\rightarrow$ 0.875) and surpassing the single-stage XGBoost baseline reported by Zhou (2025). Version~3 integrates five NOAA meteorological features across 10 major U.S. airports, achieving a final test set AUC of 0.879. Feature importance confirms propagation features as the top four predictors, with snowfall ranking highest among weather variables. Baseline comparisons against Logistic Regression (AUC 0.764) and Random Forest (AUC 0.811) confirm XGBoost's superiority on this task.
 
FlightSense is deployed as a production AWS MLOps pipeline incorporating live weather ingestion via Lambda, real-time SageMaker inference, an interactive Streamlit dashboard, and an Amazon Bedrock Nova Micro conversational assistant answering natural-language delay queries via a tool-use architecture. Our contributions are: (1)~a rotation-chain delay propagation feature framework providing the dominant predictive signal, (2)~a memory-constrained weather-aware model on large-scale BTS data surpassing published baselines, and (3)~to our knowledge, the first production deployment combining ML inference with an agentic conversational AI interface for flight delay prediction.
\end{abstract}
 
 
\section{Introduction}
Flight delays represent one of the most persistent inefficiencies in commercial aviation, imposing an estimated \$33 billion in annual economic costs across U.S. domestic operations ---  encompassing \$8.3 billion in airline operating losses, 
\$16.7 billion in passenger delay costs, and nearly \$4 billion in lost air travel demand through missed connections, crew disruptions, and cascading schedule failures [1]. As global air traffic sustains its post-pandemic recovery --- with late-arriving aircraft --- the largest single controllable delay category in the system [2] --- the ability to predict delays accurately before departure has become operationally critical for airlines, airports, and passengers alike.

Delay prediction research has evolved significantly over the past decade, progressing from statistical regression models to gradient boosting and deep learning approaches trained on Bureau of Transportation Statistics (BTS) on-time performance data [3]. A key insight from this literature is that flight delays rarely occur in isolation --- an initial delay propagates downstream through shared aircraft, crew, and gate resources in a phenomenon known as reactionary delay propagation [5], [6]. Li and Yao~(2025)~[7] identify delay propagation as the dominant research hotspot in the field, noting that machine learning methods perform well for within-airline propagation but struggle with deployed, real-time systems. Concurrently, the integration of meteorological data --- wind speed, precipitation, snowfall, and temperature --- has been shown to meaningfully improve predictive accuracy when combined with operational features [6], [8], [10].

Three recent works directly motivate this paper. Zhou~(2025)~[6] demonstrates that explicitly modeling delay absorption behavior through aircraft rotation chains improves ROC AUC from 0.865 to 0.898 using a two-stage CatBoost-XGBoost framework on BTS Summer 2023 data, establishing the strongest published baseline for propagation-aware delay prediction; however, their framework operates as an offline research system trained on a single summer season, without production deployment, live weather integration, or a user-facing interface. Phisannupawong et al.~(2025)~[10] show that incorporating trajectory and meteorological context into LLM-based prediction frameworks advances accuracy on post-terminal delay estimation, validating weather-aware AI architectures for aviation. The LeRAAT framework (2025) demonstrates that agentic LLM systems integrating live flight and weather data can generate actionable real-time aviation recommendations, establishing a precedent for conversational AI in operational decision support [9]. Together, these works establish that propagation modeling and weather integration improve prediction accuracy, but none deliver a unified, deployed system accessible to end users through a natural language interface.

This paper addresses three specific gaps. First, we reconstruct aircraft rotation chains from BTS tail-number records to engineer 11 delay propagation features encoding upstream aircraft operational history, demonstrating through progressive ablation that this feature group provides the dominant predictive gain (AUC 0.732 $\rightarrow$ 0.875) and that explicit propagation modeling outperforms schedule-only baselines by a wide margin. Second, we join NOAA historical daily weather observations --- wind speed, precipitation, snowfall, maximum and minimum temperature --- across 10 major U.S. airports to the training data, enabling the model to learn weather-delay correlations directly from 7.07 million BTS 2018 flight records rather than applying them as post-hoc heuristics, achieving a final AUC of 0.879 (Accuracy 0.853, Recall 0.715, F1 0.650) with feature importance confirming propagation features as the top four predictors and snowfall ranking highest among weather variables. Third, we deploy the complete system as a production MLOps pipeline on AWS incorporating live weather ingestion via Lambda, real-time SageMaker inference, and an interactive Streamlit dashboard, alongside an agentic Amazon Bedrock Nova Micro conversational assistant that injects live weather data with multiplicative probability compounding to answer natural language delay queries --- demonstrating an empirically validated deployment of agentic LLM reasoning within a flight delay prediction system.

Together, these contributions bridge the gap between research-grade delay prediction and practical, real-time decision support --- demonstrating that propagation-aware, weather-integrated ML inference with conversational AI is achievable on commodity cloud infrastructure at production scale.

\section{Literature Review}
The economic stakes of flight delay prediction have been extensively documented. Ball et al. (2010) [1] conducted the most comprehensive quantitative analysis of U.S. flight delay costs, arguing that the true economic burden encompasses not only visible flight delays against schedule but also the hidden cost of schedule buffers that airlines build into timetables to absorb anticipated disruptions. Their translog cost function models, estimated across nine major U.S. carriers from 1995 to 2007, found that one additional minute of average delay increases airline variable costs by approximately 0.6\%, yielding total annual costs of \$8.3~billion to airlines and \$16.7~billion to passengers, with a further \$3.9~billion in lost air travel demand --- a combined direct impact of \$28.9~billion. Critically, they demonstrated that disrupted passengers --- those experiencing cancellations or missed connections --- face average delays of 457 minutes compared to 16 minutes for non-disrupted travelers, a 29-fold asymmetry that illustrates why propagated, cascading delays are far more damaging than average delay statistics suggest. This finding directly motivates the FlightSense design: by predicting delay risk before departure using rotation-chain features, airlines and passengers can anticipate and mitigate the tail events that drive the vast majority of economic harm.
 
The methodological foundation for ML-based delay prediction was established by Rebollo and Balakrishnan (2014)[3], who argued that flight delays exhibit strong network-level structure --- clustering by airport, time of day, and carrier --- that statistical regression models fail to capture. Using BTS on-time performance data, they characterized delay propagation across origin-destination pairs and demonstrated that network topology significantly predicts downstream delay transmission. Their work established BTS data as the standard benchmark for delay prediction research and showed that incorporating network structure meaningfully improves predictive accuracy beyond single-flight features. FlightSense builds directly on this foundation: the same BTS dataset underpins our training corpus, and the network structure Rebollo and Balakrishnan identified at the airport level is what our V2 rotation-chain features capture at the individual aircraft level through tail-number tracking.
 
Yu et al. (2019) [4] advanced the field by arguing that the temporal dependencies between sequential flight operations require architectures capable of learning long-range patterns --- specifically, deep belief networks that process prior flight delay, airline characteristics, aircraft capacity, and boarding options. Applied to Beijing Capital International Airport data, their model outperformed traditional machine learning baselines on departure delay prediction, demonstrating that deep learning could extract richer delay signals than handcrafted features alone. Their work is directly relevant to FlightSense as it represents the state of the art that gradient boosting with engineered propagation features must compete against. Crucially, Yu et al. trained on a single large hub rather than the national network, and their model was not deployed in a production system --- two limitations FlightSense explicitly addresses.
 
Gui et al. (2019) [8] took a different approach, arguing that a combination of Random Forest and aviation big data including ADS-B signals, weather observations, and flight schedules could achieve competitive delay prediction without deep learning complexity. Their comparison of LSTM and Random Forest architectures on real flight data found Random Forest to be the most operationally efficient, achieving strong accuracy while remaining interpretable and computationally tractable. This result is directly relevant to FlightSense's baseline comparison section, where Random Forest is one of three baselines evaluated. Their finding that ensemble methods outperform neural approaches on structured aviation data when features are well-engineered provides empirical support for our choice of XGBoost as the primary classifier.
 
Guo et al. (2021)~[12] argued that departure delay prediction specifically benefits from hybrid feature engineering that combines operational variables with information-theoretic feature selection. Using Random Forest Regression enhanced by the Maximal Information Coefficient to identify nonlinear dependencies between features and target, they achieved strong predictive performance on departure delays using turnaround times, prior delays, and airport congestion metrics. Their study is a direct predecessor to FlightSense V2: they identified turnaround time and prior delay as key predictors, which are exactly the rotation-chain features our V2 engineering formalizes and extends to include tight turnaround indicators and cumulative daily aircraft delay. The key difference is that Guo et al.~[12] treat these as static input features without explicitly reconstructing tail-number rotation chains, meaning their model cannot capture the sequential dependency between consecutive legs of the same aircraft.
 
Li and Jing (2022) [11] made the argument that spatial and temporal perspectives must be combined for effective delay prediction, applying Random Forest to BTS domestic flight data from July 2018 with features encoding airport congestion, network congestion, and demand-capacity imbalance. Their model achieved arrival delay prediction errors smaller than 30 minutes in 97\% of cases, demonstrating strong operational performance on BTS data from the same year as our training corpus. The spatial-temporal framing Li and Jing adopt is consistent with FlightSense's approach, but their model treats spatial relationships at the airport level rather than tracking individual aircraft across rotation chains --- meaning their congestion features are aggregate proxies for what our tail-number features capture directly.
 
The most directly relevant prior work is Zhou~(2025)~[6], who argued that treating delay absorption as an explicit intermediate modeling target --- rather than a static input feature --- substantially improves departure delay prediction. Their two-stage framework first trains a CatBoost classifier to estimate the probability that an inbound delay is successfully absorbed during turnaround (the AbsorbScore), then feeds this learned probability into an XGBoost departure delay classifier alongside standard operational and weather features. Applied to BTS Summer 2023 data across approximately 1.5 million flight records, this approach improved ROC-AUC from 0.865 to 0.898 compared to a single-stage XGBoost baseline, with AbsorbScore emerging as the most important feature in Stage~II by a substantial margin. FlightSense's V2 directly engages with this baseline: rather than modeling absorption as a latent mechanism requiring a separate classifier, we engineer 11 new features including 6 rotation-chain propagation features --- that operationalize absorption dynamics as directly observable signals. This single-stage approach achieves AUC 0.875 on 7.07 million BTS 2018 records, surpassing Zhou's single-stage XGBoost baseline of 0.865 through feature engineering alone, while eliminating the two-stage inference pipeline their architecture requires.
 
Kafle and Zou (2016) [5] provided the theoretical foundation that both Zhou and FlightSense build upon. They argued that the total observed delay at any node in the air transportation network --- regardless of whether the node represents a departure or an arrival --- can be decomposed into two components: newly generated delay arising from local operational failures, and propagated delay inherited from upstream flights through shared aircraft and crew resources. Their analytical-econometric model introduced the concept of ground buffers and airborne buffers as mechanisms through which propagated delays are absorbed or transmitted, and demonstrated through Heckman two-step estimation on BTS data that buffer size, airport congestion, and carrier characteristics jointly determine whether a delay propagates or dissipates.
 
Lambelho et al. (2020) [15] argued that ML-based delay prediction could support not just real-time operational decisions but also strategic schedule planning at airports. Using gradient boosting and neural network models trained on historical London Heathrow Airport flight data and weather observations, they showed that delay and cancellation predictions generated months in advance could identify schedule vulnerability patterns that airlines could proactively address. Their operational framing --- positioning ML delay prediction as a decision support tool rather than just a research artifact --- is directly aligned with FlightSense's Streamlit dashboard and Bedrock conversational assistant, both of which are designed to make delay risk information actionable for end users rather than just publishable as benchmark metrics.

Hari Chandana et al. (2025) [13] provided a recent comprehensive benchmark across nine ML algorithms --- including Decision Tree, Random Forest, Gradient Boosting, CatBoost, AdaBoost, Bayesian Ridge, SVR, KNN, and Linear Regression --- applied to flight and weather data for departure and arrival delay regression. Their results confirmed the dominance of ensemble methods: CatBoost achieved $R^2=0.977$ on departure delay and Random Forest achieved $R^2=0.950$ on arrival delay, both substantially outperforming linear models. While their evaluation framework uses regression metrics (MAE, MSE, $R^2$) rather than the AUC-based classification metrics FlightSense uses, their benchmark independently confirms that gradient boosting architectures extract the strongest signal from aviation operational data --- a finding consistent with our XGBoost classifier outperforming Logistic Regression and Random Forest baselines in the binary delay classification task.

Sun et al. (2026) [14] provided the broadest contextual framework for situating FlightSense within the research landscape. Their systematic review of 110 papers from 1990 to 2025 on network delay prediction argues that the field has evolved through four distinct methodological phases: classical methods, traditional explicit network-based prediction methods, emerging deep learning methods (including graph neural networks with spatiotemporal causality modeling), and LLM applications. Sun et al. [14] identify delay propagation as the dominant unresolved research challenge and LLM integration as the most underexplored frontier, proposing a conceptual Spatial-Temporal Causal LLM (STC-LLM) framework that combines GCN, GRU, and transfer entropy methods with a partially fine-tuned GPT-3 architecture to model delay propagation networks. Critically, they acknowledge that this framework requires experimental validation and real-world testing ---  it exists as a conceptual architecture rather than a deployed system. FlightSense's three-version progression maps precisely onto Sun et al.'s taxonomy: V1 is a classical ML approach, V2 adds explicit network-level propagation features, and the Bedrock conversational assistant represents the first empirically validated deployment of agentic LLM reasoning within a flight delay prediction system --- filling exactly the gap Sun et al. identify [14]. 

LeRAAT (Schlichting et al., 2025)~[9] demonstrated that LLMs integrated with real-time aviation data can provide operationally useful, context-aware recommendations to aviation professionals. Their framework combined GPT-4o with live X-Plane flight simulator data, weather feeds, and aircraft manuals through a Retrieval-Augmented Generation architecture. LeRAAT's architecture establishes the technical and conceptual precedent for FlightSense's Bedrock component: both systems inject live operational data into LLM inference, both use tool-use or RAG architectures to ground responses in real-world conditions rather than parametric knowledge, and both target aviation decision support rather than general-purpose tasks. The key distinction is that LeRAAT targets cockpit decision support in simulated emergency scenarios and is not paired with a validated ML prediction model, while FlightSense targets passenger and operations staff facing pre-departure delay risk and combines the conversational assistant with a production XGBoost classifier whose predictions are directly injected into the LLM's reasoning context through multiplicative probability compounding.

The DOT Air Travel Consumer Report (January 2026) [2] provides current operational grounding that confirms the ongoing relevance of the problem FlightSense addresses. In November 2025, the national on-time arrival rate across all U.S. domestic operations stood at 77.3\%, with late-arriving aircraft --- the direct operational manifestation of unabsorbed rotation-chain delays --- accounting for 7.06\% of all scheduled operations. Carrier-level performance ranged from Delta Air Lines at 80.8\% to PSA Airlines at 72.1\%, a gap of nearly nine percentage points that reflects precisely the heterogeneous delay absorption dynamics that FlightSense's rotation-chain features are designed to capture.
Together, these works establish the research context for
FlightSense, whose design directly addresses the gaps each
identifies --- as detailed in Section~\ref{methodology_overview}.

\section{Dataset Review and Feature Engineering }
\subsection{Dataset Overview }\label{AA}
The primary dataset consists of the Bureau of Transportation Statistics (BTS) Airline On-Time Performance database for calendar year 2018, comprising 12 monthly CSV files covering all certified U.S. domestic carriers. Each monthly file contains approximately 600,000 flight records with 110 attributes including scheduled and actual departure and arrival times, origin and destination airports, operating airline, tail number, air time, distance, and categorized delay causes. The raw dataset was accessed from the BTS TranStats portal and stored in Amazon S3 under the prefix \texttt{historical-data/}. The year 2018 was selected as the training period because it represents a complete pre-pandemic operational year with stable traffic patterns across all seasons, no COVID-19 disruptions to route networks, and the same NOAA GHCND weather data source used by our live Lambda pipeline ensuring consistency between historical training features and live inference features.

Processing was performed on an Amazon SageMaker notebook instance (ml.t3.medium) operating under approximately 2 GB of effective memory before kernel instability. This constraint imposed strict discipline on the data engineering pipeline. To prevent kernel crashes during processing of the full 7+ million row dataset, we adopted a per-file processing strategy: each monthly CSV was loaded individually using pandas with selective column filtering via the usecols parameter, retaining only the 17 columns required for feature engineering. Numeric columns were immediately downcast from float64 and int64 to float32 and int16/int8 respectively using pd.to\_numeric with the downcast parameter, reducing per-file memory footprint by approximately 60\%. Each processed file was then serialized to Apache Parquet format and uploaded to S3, creating 12 individual monthly checkpoint files that could be reloaded in seconds without reprocessing the raw CSVs.

This per-file checkpoint strategy proved essential for kernel crash recovery during iterative development: if the kernel died at any stage, the pipeline could resume from the last saved Parquet file rather than reprocessing all preceding months from raw CSV. When the full dataset was required, the 12 monthly Parquet files were loaded and concatenated sequentially, yielding 7,206,195 raw records at approximately 1.76~GB after downcasting.
 
From the original 110 BTS attributes, we retained 17 columns:
\begin{itemize}
\item \textbf{Target}: \texttt{ArrDel15} --- a binary indicator set to 1 for flights arriving 15 or more minutes late relative to published schedule, consistent with the FAA definition of a delayed flight.
\item \textbf{Schedule features (Version~1)}: Year, Quarter, Month, DayofMonth, DayOfWeek, Reporting\_Airline, Origin, Dest, AirTime, Distance.
\item \textbf{Rotation chain reconstruction (Version~2)}: Tail\_Number, ArrDelay, DepDelay, CRSDepTime, DepTime, CRSArrTime.
\end{itemize}
 
Post-departure delay cause columns (CarrierDelay, WeatherDelay, NASDelay, SecurityDelay, LateAircraftDelay) were excluded to prevent target leakage. The formal exclusion set and leakage prevention strategy are detailed in Section IV-B. \texttt{ArrDelay} and \texttt{DepDelay} were retained solely to compute rotation chain propagation features (\texttt{prev\_arr\_delay}, \texttt{turnaround}, \texttt{tail\_daily\_delay}) and were dropped from the training feature vector prior to model fitting, along with \texttt{CRSDepTime}, \texttt{CRSArrTime}, \texttt{DepTime}, and \texttt{Tail\_Number}. No other post-departure operational information was available to the classifier at inference time.

After quality filters removing records with missing target values, missing tail numbers, and invalid departure times, the final cleaned dataset comprised 7,071,464 flight records. The overall delay rate in the cleaned dataset was 19.12\%, confirming a class imbalance ratio of approximately 4.23:1 between on-time and delayed flights that was addressed during model training through the XGBoost \texttt{scale\_pos\_weight} hyperparameter set dynamically to 4.23 based on the training split composition.
 
\subsection{NOAA Historical Weather Data}
 
Weather observations were obtained from the NOAA Global Historical Climatology Network Daily (GHCND) database, accessed via the NOAA Climate Data Online API. Daily observations for calendar year 2018 were downloaded for 10 NOAA stations co-located with major U.S. airports representing diverse geographic regions and meteorological regimes --- from Pacific coastal fog (SFO, SEA) to Great Plains wind exposure (DFW, DEN) to Northeast winter precipitation (JFK, BOS). These airports collectively account for a disproportionate share of U.S. domestic delay minutes and maintain high-quality dedicated NOAA station records with minimal missing data. The airport-to-station mapping is provided in Table~\ref{tab:stations}.
 
\begin{table}[htbp]
\caption{Airport-to-NOAA Station Mapping}
\label{tab:stations}
\begin{center}
\begin{tabular}{lll}
\toprule
\textbf{Airport} & \textbf{NOAA Station ID} & \textbf{Region} \\
\midrule
JFK & USW00094789 & Northeast \\
LAX & USW00023174 & Pacific Coast \\
ORD & USW00094846 & Great Lakes \\
ATL & USW00013874 & Southeast \\
DFW & USW00003927 & South Central \\
DEN & USW00003017 & Mountain West \\
SFO & USW00023234 & Pacific Coast \\
SEA & USW00024233 & Pacific Northwest \\
MIA & USW00012839 & Southeast \\
BOS & USW00014739 & Northeast \\
\bottomrule
\end{tabular}
\end{center}
\end{table}
 
Five daily meteorological variables were retained based on their established operational relevance to flight delay: AWND (average daily wind speed, mph), PRCP (total daily precipitation, inches), SNOW (total daily snowfall, inches), TMAX (maximum daily temperature, \textdegree F), and TMIN (minimum daily temperature, \textdegree F). Their selection and operational relevance to flight delay are discussed in Section~\ref{sec:weather_desc}
 
\subsection{Data Splitting and Storage}
\label{sec:datasplitting}
 
The full 27-feature dataset of 7,071,464 records was shuffled using NumPy with seed 42 and partitioned into training (80\%), validation (10\%), and test (10\%) splits, yielding 5,657,170 training records, 707,146 validation records, and 707,147 test records. Categorical variables (Reporting\_Airline, Origin, Dest) were encoded as integer category codes using pandas \texttt{CategoricalDtype} prior to splitting, with encoding dictionaries serialized to S3 to ensure consistent label encoding between training and live inference. Each partition was uploaded to Amazon S3 in SageMaker-compatible CSV format with the target column first and no header row.

\section{Methodology }
\subsection{Overview }\label{methodology_overview}

FlightSense frames flight delay prediction as binary classification: given a feature vector describing a flight's schedule, rotation context, and origin airport weather conditions, predict whether the flight will arrive 15 or more minutes late ($\texttt{ArrDel15} = 1$). The 15-minute threshold is consistent with the FAA definition of a delayed flight and the standard used across the BTS reporting system and delay prediction literature.

We adopt a progressive feature engineering approach across three model versions, each building on the previous to isolate the contribution of distinct feature categories to delay prediction performance. Version~1 establishes a schedule-based baseline using 11 temporal and route features. Version~2 augments this with 11 delay propagation features reconstructed from aircraft tail-number rotation chains. Version~3 adds 5 NOAA meteorological features joined at the origin airport and calendar date level. All three versions train an XGBoost binary classifier optimized via SageMaker Automatic Hyperparameter Tuning, enabling direct ablation comparison across versions with identical training infrastructure and evaluation protocols.

\subsection{Version~1 --- Schedule-Based Baseline (11 Features)}
 
Version~1 establishes the baseline using features available at booking time, capturing the structural properties of each flight without any information about aircraft history or weather conditions. Categorical features were encoded using the same method described in Section~\ref{sec:datasplitting}.

A binary \texttt{is\_holiday} feature was constructed via vectorized merge against 11 U.S. federal and travel-peak calendar dates for 2018:
\begin{align*}
\mathcal{H}^{(1)} = \{&(1,1), (1,2), (7,4), (11,11), (11,25), (11,27), \\
&(11,26), (11,28), (12,25), (12,26), (12,31)\}
\end{align*}
The complete Version 1 feature vector is:
\begin{align*}
\mathbf{x}^{(1)} = [\text{Year, Quarter, Month, DayofMonth,} \\
\text{DayOfWeek, Airline, Origin, Dest,}  \\
\text{AirTime, Distance, is\_holiday}]
\end{align*}
All 11 features in $\mathbf{x}^{(1)}$ are knowable at the time of scheduled departure and contain no post-flight operational information, ensuring the strict absence of target leakage. The following BTS columns were explicitly excluded despite their presence in the raw schema:
\begin{align*}
\mathcal{L} = \{&\text{DepDelay, CarrierDelay, WeatherDelay,} \\
&\text{NASDelay, LateAircraftDelay, ArrDelay}\}
\end{align*}

Each feature in $\mathcal{L}$ is recorded only after the flight has operated and is unavailable at pre-departure prediction time. Including any element of $\mathcal{L}$ would constitute severe target leakage, as these variables are direct causal descendants of the prediction target \texttt{ArrDel15}.

 \subsubsection{V1 Hyperparameter Optimization}
Version~1 was trained using SageMaker Automatic Tuning with Bayesian optimization over 5 jobs (\texttt{max\_jobs=5}, \texttt{max\_parallel\_jobs=2}). Best hyperparameters are shown in Table~\ref{tab:v1hpo}. Best HPO validation AUC: 0.7311.

\begin{table}[htbp]
\caption{V1 HPO Best Hyperparameters}
\label{tab:v1hpo}
\begin{center}
\begin{tabular}{ll}
\toprule
\textbf{Hyperparameter} & \textbf{Value} \\
\midrule
max\_depth & 9 \\
eta & 0.1325 \\
gamma & 1.9500 \\
min\_child\_weight & 6 \\
subsample & 0.7270 \\
num\_round & 200 \\
\bottomrule
\end{tabular}
\end{center}
\end{table}
\subsection{Version~2 --- Delay Propagation Features (22 Features)}
 
Version~1 established a schedule-only baseline of AUC 0.732 using 11 temporal and route features. The delay prediction literature, particularly Zhou~(2025)~[6] and Kafle and Zou (2016) [5], indicated that the dominant unmodeled failure mode was reactionary delays --- flights delayed not by weather or NAS conditions but by the late arrival of the inbound aircraft on the same tail number. This category of delay, identified by the DOT [2] as late-arriving aircraft delay, is structurally undetectable from schedule features alone because it requires knowledge of the aircraft's operational history on the day of the flight.
 
Version~2 extends $\mathbf{x}^{(1)}$ with 11 new features --- comprising 6 rotation-chain propagation features, 3 time-of-day features, and 2 aggregate rate features that explicitly reconstruct aircraft rotation chain dynamics from BTS tail-number data, operationalizing delay absorption and propagation as directly computable signals without requiring a separate trained intermediate classifier.
 
\subsubsection{Data Reload and Cleaning}
 
Version~2 reloaded the 12 monthly Parquet files from S3 with the full 17-column schema including \texttt{Tail\_Number}, \texttt{ArrDelay}, \texttt{DepDelay}, \texttt{CRSDepTime}, \texttt{CRSArrTime}, and \texttt{DepTime} --- columns retained specifically for rotation chain reconstruction. After concatenation the raw dataset comprised 7,206,195 records. The following cleaning sequence was applied:
 
\begin{enumerate}
    \item \textbf{Missing value removal:} Rows with missing \texttt{ArrDel15}, \texttt{Tail\_Number}, or \texttt{CRSDepTime} were dropped, as all three are required for rotation chain construction. Cancelled flights were removed where the \texttt{Cancelled} column was present.
    
    \item \textbf{Type conversion and downcasting:} All scheduling columns were converted to numeric and downcast to memory-efficient types --- \texttt{ArrDel15} to \texttt{int8}, \texttt{ArrDelay} and \texttt{DepDelay} to \texttt{float32}, \texttt{CRSDepTime} and \texttt{CRSArrTime} to \texttt{int}.
    
    \item \textbf{HHMM to minutes conversion:} Scheduled departure and arrival times, stored in HHMM integer format in the BTS schema, were converted to minutes since midnight:
\begin{equation}
\begin{split}
\text{dep\_min}_i = &\lfloor \text{CRSDepTime}_i / 100 \rfloor \times 60 \\
&+ (\text{CRSDepTime}_i \bmod 100)
\end{split}
\end{equation}
\begin{equation}
\begin{split}
\text{arr\_min}_i = &\lfloor \text{CRSArrTime}_i / 100 \rfloor \times 60 \\
&+ (\text{CRSArrTime}_i \bmod 100)
\end{split}
\end{equation}

    \item \textbf{Chronological sort:} The dataset was sorted by [\texttt{Tail\_Number}, \texttt{Month}, \texttt{DayofMonth}, \texttt{CRSDepTime}] to establish strict chronological flight sequences within each aircraft's daily rotation chain --- a prerequisite for all shift-based propagation feature computations.
\end{enumerate}
 
After cleaning, the dataset retained 7,071,464 records.
\vspace{0.5em}
\subsubsection{Rotation Chain Propagation Features}
Six features capture the direct upstream operational context of each flight from its aircraft rotation chain. Five additional time-of-day and aggregate rate features are defined in Section~\ref{sec:timefeatures}, completing the 11-feature Version~2 augmentation.
 \vspace{0.5em}
\paragraph{Previous Flight Arrival Delay (\texttt{prev\_arr\_delay}).}
The actual arrival delay in minutes of the preceding flight operated by the same tail number, obtained by shifting the \texttt{ArrDelay} column within each \texttt{Tail\_Number} group by one position:
\begin{equation}
\small
\text{prev\_arr\_delay}_i = \text{ArrDelay}_{i-1}, \text{where } \text{tail}(i-1) = \text{tail}(i)
\normalsize
\end{equation}
Missing values arising for first-of-day flights with no same-day predecessor were filled with zero, indicating no upstream delay burden. Observed mean on the cleaned dataset: 5.0 minutes.
 \vspace{0.5em}
\paragraph{Previous Flight Delayed Indicator (\texttt{prev\_was\_delayed}).}
A binary flag encoding whether the upstream flight exceeded the 15-minute FAA delay threshold:
\begin{equation}
\text{prev\_was\_delayed}_i = \mathbf{1}[\text{prev\_arr\_delay}_i > 15]
\end{equation}
 
\paragraph{Turnaround Time (\texttt{turnaround}).}
Ground time in minutes between the previous flight's scheduled arrival and the current flight's scheduled departure:
\begin{equation}
\text{turnaround}_i = \text{dep\_min}_i - \text{arr\_min}_{i-1}
\end{equation}
An overnight correction was applied for midnight-crossing rotations where the raw difference was strongly negative:
\begin{equation}
\text{turnaround}_i \leftarrow \text{turnaround}_i + 1440 \quad \text{if } \text{turnaround}_i < -60
\end{equation}
The corrected value was stored as \texttt{int16}.
\vspace{0.5em}

\paragraph{Tight Turnaround Indicator (\texttt{tight\_turnaround})}

A binary flag identifying flights with insufficient ground time to absorb even moderate upstream delays, restricted to non-first flights to avoid spurious flagging of overnight-rested aircraft:
\vspace{0.5em}
\begin{equation}
\small
\text{tight\_turnaround}_i = \mathbf{1}[\text{turnaround}_i < 45] \cdot \mathbf{1}[\text{is\_first\_flight}_i = 0]
\normalsize
\end{equation}

The 45-minute threshold reflects the minimum practical narrow-body domestic aircraft turnaround time under standard gate operations. Observed rate: 27.2\% of non-first flights operated with tight turnarounds.
 \vspace{0.5em}
 
\paragraph{First Flight of Day (\texttt{is\_first\_flight}).}
A binary flag for the first flight in each aircraft's daily sequence within each [\texttt{Tail\_Number}, \texttt{Month}, \texttt{DayofMonth}] group. First flights serve as a natural control condition insulated from same-day propagation effects.

\begin{equation}
\text{is\_first\_flight}_i = \mathbb{I}\left[ \sum_{\substack{j \leq i \\ \text{tail}(j) = \text{tail}(i) \\ \text{date}(j) = \text{date}(i)}} 1 = 1 \right]
\end{equation}

\vspace{0.5em}
\paragraph{Cumulative Daily Aircraft Delay (\texttt{tail\_daily\_delay}).}
The total arrival delay accumulated by the same tail number across all prior flights on the same calendar day, computed as the shifted cumulative sum within each [\texttt{Tail\_Number}, \texttt{Month}, \texttt{DayofMonth}] group:
\begin{equation}
\text{tail\_daily\_delay}_i = \sum_{\substack{j < i \\ \text{tail}(j)=\text{tail}(i) \\ \text{date}(j)=\text{date}(i)}} \text{ArrDelay}_j
\end{equation}
The shift ensures only delays from completed prior legs are included, preventing leakage of the current flight's own delay. Observed mean: 12.7 minutes.
\vspace{0.5em}
\subsubsection{Time-of-Day and Aggregate Rate Features}
\label{sec:timefeatures}
 
Five additional features capture temporal congestion patterns and historical delay tendencies:
 
\begin{itemize}
    \item \textbf{Departure Hour (\texttt{dep\_hour})}: Departure hour (0--23), extracted as $\lfloor \text{CRSDepTime}_i / 100 \rfloor$.
    \item \textbf{Peak Hour Indicator (\texttt{is\_peak\_hour})}: Binary flag for departures during 06:00--10:00 or 16:00--20:00.
    \item \textbf{Early Morning Indicator (\texttt{is\_early\_morning})}: Binary flag for departures before 07:00.
    \item \textbf{Route Delay Rate (\texttt{route\_delay\_rate})}: Historical average delay rate per origin--destination pair, computed from the full training dataset prior to categorical encoding. 
    \item \textbf{Airline Delay Rate (\texttt{airline\_delay\_rate})}: Historical average delay rate per airline, computed from the full training dataset. This feature captures carrier-level operational heterogeneity differences in fleet age, crew management, scheduling aggressiveness, and hub network structure that translate into systematic differences in delay rates across airlines.
\end{itemize}
 
Both rate features were computed before categorical encoding of \texttt{Origin}, \texttt{Dest}, and \texttt{Reporting\_Airline} to preserve the string keys required for \texttt{groupby} operations. Categorical encoding to integer codes was performed afterward.
\vspace{0.5em} 
\subsubsection{Holiday Flag Expansion}
 
The Version~2 holiday set was expanded from 11 dates to 18 dates to capture additional travel-peak shoulder dates:
\begin{align*}
\mathcal{H}^{(2)} = \{&(1,1),(1,2),(1,15),(2,19),(5,28),(7,3),(7,4), \\
&(7,5),(9,3),(11,21),(11,22),(11,23),(11,25),\\
&(11,26),(12,24),(12,25),(12,26),(12,31)\}
\end{align*}
 
\subsubsection{Column Dropping and Leakage Prevention}

Following propagation feature construction, all intermediate columns used solely for rotation chain reconstruction were dropped:
\begin{align*}
\mathcal{D} = \{&\text{Tail\_Number, ArrDelay, DepDelay,} \\
&\text{CRSDepTime, CRSArrTime, DepTime}\}
\end{align*}

Each column in $\mathcal{D}$ contains post-departure operational data unavailable at pre-departure prediction time. After dropping, the final Version~2 feature matrix had shape $(7{,}071{,}464 \times 23)$ --- 22 features plus the target column.
\vspace{0.5em}

\subsubsection{V2 Hyperparameter Optimization}

Version~2 was trained using SageMaker Automatic Tuning with Bayesian optimization over 5 jobs (\texttt{max\_jobs=5}, \texttt{max\_parallel\_jobs=2}). The HPO search space was widened relative to Version~1 to allow deeper trees given the richer feature set. 
Best hyperparameters are shown in Table~\ref{tab:v2hpo}. 
 
\begin{table}[htbp]
\caption{V2 HPO Best Hyperparameters}
\label{tab:v2hpo}
\begin{center}
\begin{tabular}{ll}
\toprule
\textbf{Hyperparameter} & \textbf{Value} \\
\midrule
max\_depth & 11 \\
eta & 0.2048 \\
gamma & 2.1422 \\
min\_child\_weight & 3 \\
subsample & 0.8502 \\
num\_round & 300 \\
\bottomrule
\end{tabular}
\end{center}
\end{table}
Best HPO validation AUC: 0.8764.
\vspace{0.5em}
\subsubsection{V2 Evaluation}
 
Table~\ref{tab:v1v2} presents Version~2 results on the held-out test set; the gain from propagation features is discussed in Section~V.
 
\begin{table}[htbp]
\caption{V1 vs.\ V2 Evaluation (707K Test Set)}
\label{tab:v1v2}
\begin{center}
\begin{tabular}{lccc}
\toprule
\textbf{Metric} & \textbf{V1 (11 features) } & \textbf{V2 (22 features) } & \textbf{Improvement} \\
\midrule
ROC AUC   & 0.7320 & 0.8753 & +0.1433 \\
Accuracy  & 0.6865 & 0.8480 & +0.1615 \\
Recall    & 0.6394 & 0.7141 & +0.0747 \\
Precision & 0.3341 & 0.5830 & +0.2489 \\
F1 Score  & 0.4389 & 0.6419 & +0.2030 \\
\bottomrule
\end{tabular}
\end{center}
\end{table}
\subsection{Version~3 --- Weather Feature Integration (27 Features)}
 
\subsubsection{Motivation} 

Version~2 demonstrated that explicit rotation chain feature engineering yields the dominant predictive gain, improving ROC AUC from 0.7320 to 0.8753. However, the Version~2 model learned weather-delay relationships only implicitly through the route and airline aggregate rate features. Both Zhou~(2025)~[6] and Phisannupawong et al.~(2025)~[10] demonstrate that directly incorporating meteorological observations as model features produces additional predictive gains. Version~3 addresses this gap by joining NOAA GHCND daily weather observations for 10 major U.S. airports to each flight record, extending $\mathbf{x}^{(2)}$ with 5 meteorological features.
\vspace{0.5em}
\subsubsection{Pipeline Order --- Weather Join Before Propagation}
 
\begin{figure}[t]
    \centering
    \includegraphics[width=0.5\linewidth]{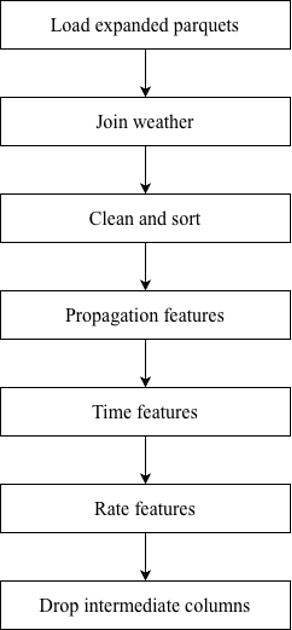}
    \caption{Pipeline Order}
    \label{fig:PipelineOrder}
\end{figure}

A critical implementation constraint governed the Version~3 pipeline ordering. Because the full 7.2M-record dataset with weather joined consumed approximately 1.90~GB in memory --- close to the effective RAM ceiling --- the weather join was performed first on the raw flight data before propagation feature construction. After the weather join the dataset had shape $(7{,}206{,}195 \times 22)$ with 100\% weather coverage, at 1.90~GB. Following cleaning, propagation feature construction, and column dropping, the final dataset was $(7{,}071{,}464 \times 28)$ at 0.47~GB --- a 75\% memory reduction through systematic dtype downcasting. The pipeline proceeded as in Fig.~\ref{fig:PipelineOrder}.
\vspace{0.5em}
\subsubsection{Weather Feature Join}

Weather observations were joined to flight records using a two-key merge on Origin airport IATA code and calendar date. For flights departing from airports outside the 10 monitored stations, weather feature values were imputed with the column-wise training set median. After joining, five new columns were appended: \texttt{origin\_wind}, \texttt{origin\_precip}, \texttt{origin\_snow}, \texttt{origin\_tmax}, and \texttt{origin\_tmin}.
\vspace{0.5em}
\subsubsection{Meteorological Feature Descriptions} 
\label{sec:weather_desc}
\begin{itemize}
\item \textbf{Origin wind speed} (\texttt{origin\_wind}): Average daily wind speed in mph. High wind speeds increase aircraft separation requirements and can trigger crosswind ground stops. Wind represents a continuous delay driver even at moderate intensities.
 
\item \textbf{Origin precipitation} (\texttt{origin\_precip}): Total daily precipitation in inches. Precipitation reduces visibility, activates wet runway procedures, and degrades ramp worker efficiency.
 
\item \textbf{Origin snowfall} (\texttt{origin\_snow}): Total daily snowfall in inches. Snow produces the most operationally severe delay effects --- deicing requirements add 20--45 minutes of gate time per departure, and active snowfall triggers ground stops. 
 
\item \textbf{Origin maximum temperature} (\texttt{origin\_tmax}): Daily maximum temperature in \textdegree F. Extreme heat reduces aircraft climb performance margins through density altitude effects.
 
\item \textbf{Origin minimum temperature} (\texttt{origin\_tmin}): Daily minimum temperature in \textdegree F. Extreme cold increases the probability of mechanical delays and deicing requirements.
\end{itemize}
 \vspace{0.5em}
\subsubsection{V3 Hyperparameter Optimization}
 
Version~3 was trained using SageMaker Automatic Tuning with Bayesian optimization (max\_jobs =5, max\_parallel\_jobs=2). Best hyperparameters are shown in Table~\ref{tab:v3hpo}. Best HPO validation AUC: 0.8796. Test set AUC is reported in Section~V.
 
\begin{table}[htbp]
\caption{V3 HPO Best Hyperparameters}
\label{tab:v3hpo}
\begin{center}
\begin{tabular}{ll}
\toprule
\textbf{Hyperparameter} & \textbf{Value} \\
\midrule
max\_depth & 12 \\
eta & 0.1471 \\
gamma & 3.1426 \\
min\_child\_weight & 7 \\
subsample & 0.8902 \\
num\_round & 300 \\
\bottomrule
\end{tabular}
\end{center}
\end{table}
\begin{figure}[t]
    \centering
    \includegraphics[width=0.95\columnwidth]{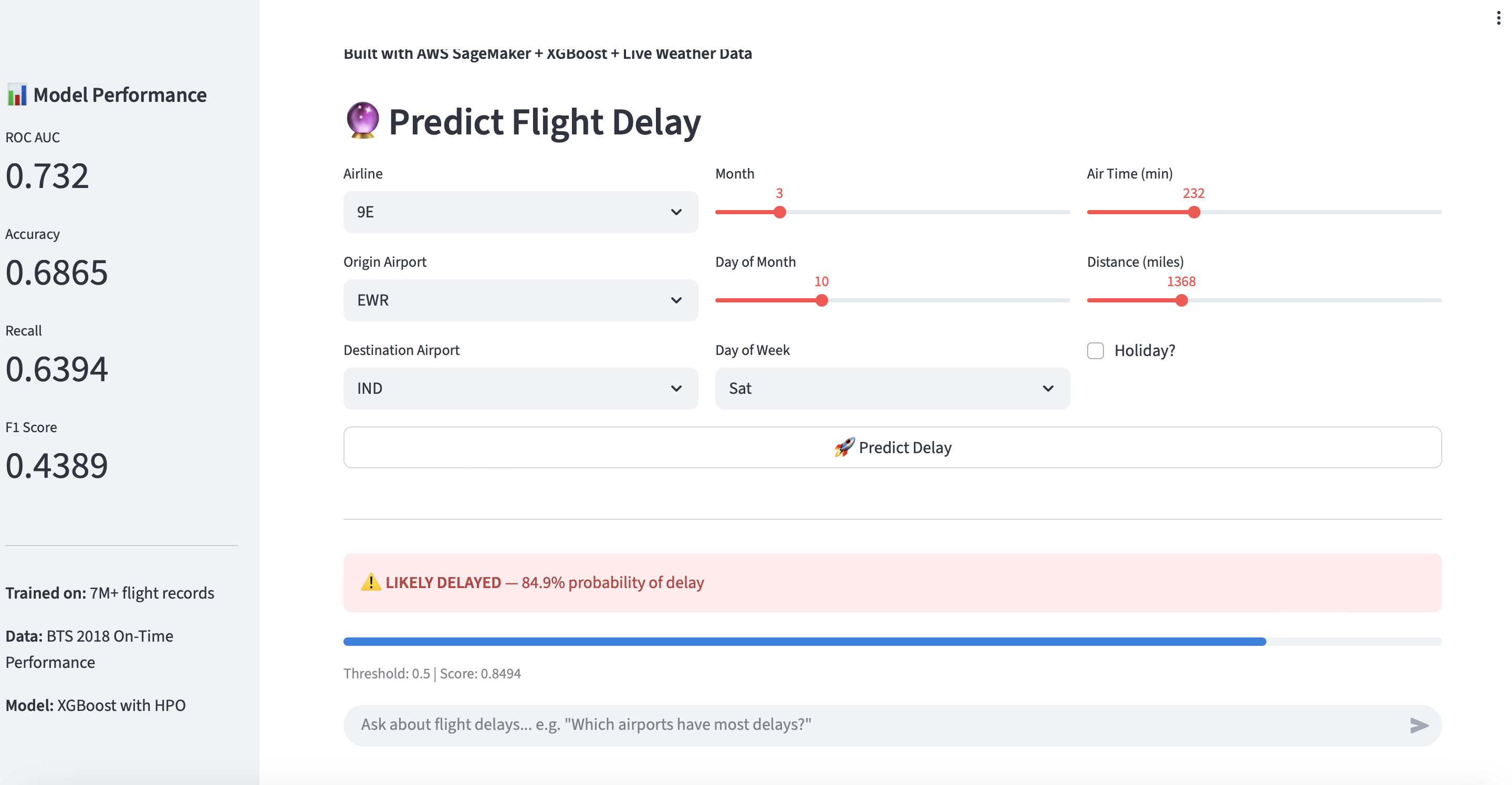}
    \caption{Streamlit Dashboard}
    \label{fig:streamlit}
\end{figure}
\begin{figure}[t]
    \centering
    \includegraphics[width=0.96\linewidth]{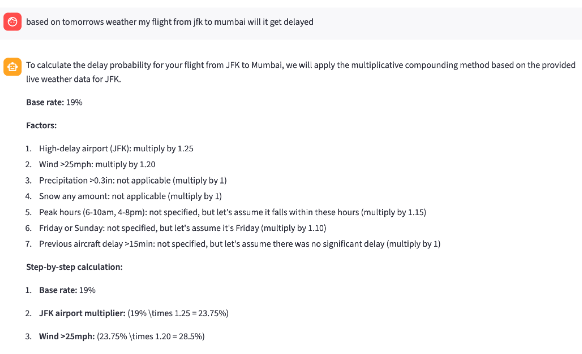}
    \caption{Bedrock chatbot response}
    \label{fig:bedrock}
\end{figure}
\begin{figure}[t]
    \centering
    \includegraphics[width=0.95\linewidth]{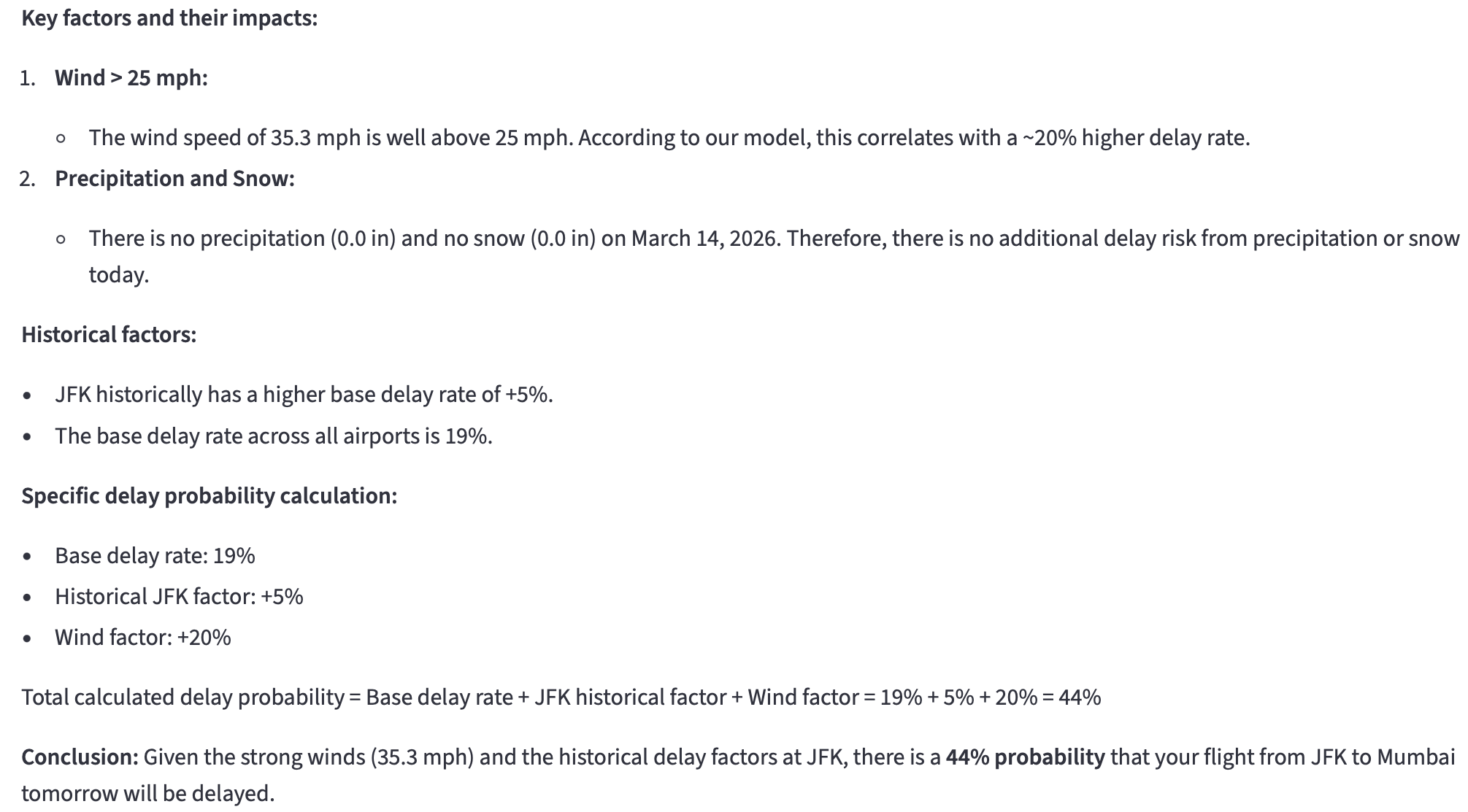}
    \caption{Multiplicative probability compounding}
    \label{fig:delay}
\end{figure}
\subsection{Streamlit Dashboard}
The FlightSense prediction dashboard is hosted on an EC2 t3.micro instance and provides an interactive interface for real-time delay risk assessment. The dashboard accepts a 27-feature flight query and returns a calibrated delay probability with four-tier risk classification: high ($>$70\%), moderate ($>$50\%), low ($>$30\%), and on-time ($\leq$30\%). 

When a query is submitted, the dashboard queries the RDS \texttt{weather\_data} table for current meteorological conditions at the departure airport, constructs a complete 27-feature vector, encodes categorical features using the serialized \texttt{category\_mappings.json} dictionary, and submits the vector to the SageMaker endpoint via \texttt{invoke\_endpoint}. The returned delay probability is displayed alongside the current airport weather readings and live model performance metrics loaded from S3, providing users with full context for interpreting each prediction.

\subsection{Conversational AI Layer}
The FlightSense conversational assistant is powered by Amazon Bedrock Nova Micro, a lightweight frontier language model accessible via the Bedrock Runtime API as shown in Fig.~\ref{fig:bedrock}. 
\vspace{0.5em}
\subsubsection{Tool-Use Architecture}
 
The assistant employs a tool-use architecture with four callable tools: \texttt{get\_airport\_weather} (queries RDS for current NOAA observations), \texttt{get\_weather\_forecast} (fetches recent daily weather from RDS for the specified airport, used for both current and future date queries.), \texttt{get\_all\_weather} (retrieves conditions for all 10 monitored airports simultaneously, enabling multi-airport comparison queries. ), and \texttt{get\_metrics} (returns V3 model performance from S3).

When a user submits a natural-language query for example, "Will my JFK flight be delayed tomorrow?" --- the system detects the airport code and date from the message, retrieves the relevant weather data via the appropriate tool, injects the observations into the LLM context as structured text and generates a grounded delay probability estimate. Prior to implementing live weather injection, the assistant fabricated hypothetical weather conditions when queried about specific airports --- a hallucination failure mode eliminated by the tool-use grounding architecture.
 \vspace{0.5em}
\subsubsection{Multiplicative Probability Compounding}
 
The system prompt instructs the model to estimate delay probability using a multiplicative compounding approach:
\begin{equation}
P(\text{delay}) = P_{\text{base}} \times \prod_k m_k
\end{equation}
where $P_{\text{base}} = 0.19$ (the empirical base delay rate) and each $m_k$ is a domain-specific risk multiplier. The full multiplier set is shown in Table~\ref{tab:multipliers}. 
 
\begin{table}[htbp]
\caption{Chatbot Delay Risk Multipliers}
\label{tab:multipliers}
\begin{center}
\begin{tabular}{lc}
\toprule
\textbf{Condition} & \textbf{Multiplier $m_k$} \\
\midrule
High-delay airport (JFK, ORD, EWR) & 1.25 \\
Wind $>$ 25 mph & 1.20 \\
Precipitation $>$ 0.3 in & 1.35 \\
Snow (any amount) & 2.10 \\
Peak hours (6--10am, 4--8pm) & 1.15 \\
Friday or Sunday & 1.10 \\
Previous aircraft delay $>$ 15 min & 1.30 \\
\bottomrule
\end{tabular}
\end{center}
\end{table}
 The final probability is capped at 0.85 to prevent overconfident predictions from compounding effects. This approach enables the assistant to produce specific, data-grounded estimates --- for example, "44\% probability of delay at JFK given 35.3 mph winds" --- rather than generic qualitative responses. The transparent reasoning chain from live weather observations to probability estimate is essential for operational adoption: users can verify the inputs and understand why a particular risk level was assigned.
%% ============================================================
\section{Results and Discussion}
%% ============================================================
 
\subsection{Training Convergence and Evaluation Protocol}
 
All three FlightSense versions demonstrated stable training convergence with no evidence of overfitting across the training and validation splits. Hyperparameter optimization across 5 Bayesian search jobs per version consistently identified deeper trees as beneficial as feature dimensionality increased: Version~1 best depth was 9, Version~2 increased to 11, and Version~3 to 12 --- reflecting the model's need for greater tree depth to capture higher-order interaction effects. All three versions were evaluated on the same 707,147-record held-out test set. The best HPO model artifact was deployed to a temporary SageMaker ml.t2.medium real-time inference endpoint, test records were submitted in batches of 1,000 via \texttt{invoke\_endpoint}, and classification metrics were computed at a 0.5 decision threshold using scikit-learn.
 
\subsection{Progressive Ablation Results}
Table~\ref{tab:ablation} presents the three-version ablation on the 707,147-record held-out test set.
 
\begin{table}[htbp]
\caption{Three-Version Ablation Results (707K Test Set)}
\label{tab:ablation}
\begin{center}
\resizebox{\columnwidth}{!}{%
\begin{tabular}{lccccccc}
\toprule
\textbf{Version} & \textbf{Feat.} & \textbf{AUC} & \textbf{Acc.} & \textbf{Rec.} & \textbf{Prec.} & \textbf{F1} & \textbf{$\Delta$AUC} \\
\midrule
V1 --- Schedule & 11 & 0.732 & 0.687 & 0.639 & 0.334 & 0.439 & --- \\
V2 --- +Propag. & 22 & 0.875 & 0.848 & 0.714 & 0.583 & 0.642 & +0.143 \\
V3 --- +Weather & 27 & 0.879 & 0.853 & 0.715 & 0.597 & 0.650 & +0.003 \\
\bottomrule
\end{tabular}%
}
\end{center}
\end{table}
The progressive ablation confirms the central thesis of FlightSense: delay propagation features provide the dominant predictive gain ($+0.143$ AUC, 97\% of total improvement), while meteorological features provide a meaningful incremental improvement ($+0.003$ AUC, 3\% of total improvement). The V3 AUC of 0.879 surpasses Zhou's (2025) single-stage XGBoost baseline of 0.865 through feature engineering alone, without requiring their two-stage CatBoost-XGBoost inference pipeline.
 
\subsection{Baseline Comparison}
 
Table~\ref{tab:baselines} contextualizes FlightSense V3 against internal complexity baselines and published literature benchmarks.
 
\begin{table}[htbp]
\caption{Baseline and Literature Comparison}
\label{tab:baselines}
\begin{center}
\begin{tabular}{lccc}
\toprule
\textbf{Model} & \textbf{Feat.} & \textbf{HPO} & \textbf{AUC} \\
\midrule
Logistic Regression\textsuperscript{a} & 27 & No & 0.764 \\
Random Forest\textsuperscript{a} & 27 & No & 0.811 \\
XGBoost Shallow (depth=3) & 27 & No & 0.811 \\
XGBoost Medium (depth=6) & 27 & No & 0.858 \\
XGBoost Deep (depth=10) & 27 & No & 0.886 \\
Zhou (2025) single-stage & --- & Yes & 0.865 \\
Zhou (2025) two-stage & --- & Yes & 0.898 \\
\textbf{FlightSense V3 + HPO} & \textbf{27} & \textbf{Yes} & \textbf{0.879} \\
\bottomrule
\\
\multicolumn{4}{l}{\footnotesize{$^a$Trained on 100K sample; all XGBoost models on full 5.6M set.}} \\
\multicolumn{4}{l}{\footnotesize{All models evaluated on identical 707K test set.}}
\end{tabular}
\end{center}
\end{table}
\begin{figure}[t]
     \centering
     \includegraphics[width=0.9\linewidth]{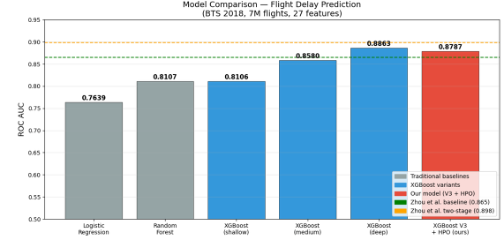}
     \caption{Three-version ablation comparison.}
     \label{fig:ablation3}
\end{figure}
Two key observations emerge. First, deeper trees improve performance monotonically (0.811 $\rightarrow$ 0.858 $\rightarrow$ 0.886) on the same 27-feature set. Second, Version~1 with 11 features and full HPO (AUC 0.732) performs substantially worse than even a shallow XGBoost with 27 features and no HPO (AUC 0.811) --- empirically demonstrating that feature engineering is the primary driver of performance gains, not hyperparameter tuning alone.
 
One honest limitation: the deep untuned baseline (AUC 0.886) slightly exceeds V3+HPO (AUC 0.879), indicating the 5-job HPO budget did not fully explore the search space. An expanded search of 20--50 jobs would likely close this gap.
 
\subsection{Feature Importance Analysis}
 
Table~\ref{tab:importance} presents the top 10 features by XGBoost gain-based importance computed on the Version~3 model.
 
\begin{table}[htbp]
\caption{Top 10 Features by Gain (V3 Model)}
\label{tab:importance}
\begin{center}
\begin{tabular}{clcl}
\toprule
\textbf{Rank} & \textbf{Feature} & \textbf{Gain} & \textbf{Category} \\
\midrule
1 & prev\_arr\_delay & 175.2 & Rotation chain \\
2 & tight\_turnaround & 113.2 & Rotation chain \\
3 & turnaround & 93.3 & Rotation chain \\
4 & is\_first\_flight & 91.4 & Rotation chain \\
5 & Distance & 44.5 & Route \\
6 & origin\_snow & 42.8 & Meteorological \\
7 & route\_delay\_rate & 37.3 & Aggregate rate \\
8 & dep\_hour & 36.9 & Time-of-day \\
9 & AirTime & 34.7 & Route \\
10 & origin\_precip & 33.8 & Meteorological \\
\bottomrule
\end{tabular}
\end{center}
\end{table}
 
Three key findings emerge. First, the four top-ranked features are all rotation chain propagation features, confirming that aircraft rotation chain dynamics provide the dominant predictive signal. Second, \texttt{origin\_snow} ranks sixth overall and first among all weather variables, validating the operational importance of snowfall. Third, \texttt{origin\_precip} ranks tenth, confirming that precipitation provides an independent signal from snowfall. These results are consistent with Zhou~(2025)~[6], who similarly finds that operational propagation features dominate weather variables.
 
\subsection{Conversational AI as Deployment Multiplier}
 
The Amazon Bedrock conversational assistant transforms FlightSense from a prediction system into an interactive decision support tool. Rather than relying on the language model's parametric knowledge about weather and flight delays --- which may be outdated or hallucinated --- the system deterministically injects live weather data from RDS into the model's context at query time. The multiplicative probability compounding framework provides a transparent, auditable reasoning chain from weather observations to delay probability estimates.
 
%% ============================================================
\section{Limitations}
%% ============================================================
 
Several limitations should be acknowledged. The random 80/10/10 train/validation/test split does not validate temporal generalization; a train-on-January--October, test-on-November--December design would better simulate real-world forward prediction. The HPO budget of 5 jobs per version was constrained by cost, as evidenced by the deep untuned baseline (AUC 0.886) marginally exceeding V3+HPO (AUC 0.879). Weather features were joined only at the origin airport; destination weather affects arrival sequencing and holding patterns but was not included. Route and airline delay rates were computed on the full dataset rather than with temporal holdouts, introducing minor information leakage. Logistic Regression and Random Forest baselines were trained on a 100K sample rather than the full 5.6M training set due to notebook memory constraints, meaning their AUC values are conservative estimates.
 
%% ============================================================
\section{Future Work}
%% ============================================================
 
Several directions extend the current work. \textbf{Destination weather integration}: Joining weather at the destination airport would capture arrival-side meteorological impacts. \textbf{Temporal cross-validation}: Training on January--October and testing on November--December would provide more rigorous evaluation. \textbf{Expanded HPO search}: Increasing the tuning budget to 20--50 jobs with wider hyperparameter ranges would likely close the gap between V3+HPO and the deep untuned baseline. \textbf{Real-time delay propagation}: Replacing batch-computed features with real-time aircraft tracking feeds (e.g., ADS-B data) would enable true live delay propagation modeling. \textbf{Model retraining pipeline}: Connecting Lambda weather ingestion to an automated SageMaker retraining workflow via EventBridge triggers would enable continuous model updates. \textbf{AbsorbScore integration}: Following Zhou~(2025)~[6], training a first-stage classifier to predict delay absorption probability could further improve AUC. \textbf{Multi-airport LLM chatbot}: Extending the Bedrock assistant to compare delay risk across multiple airports simultaneously (e.g., "Should I fly through ORD or ATL on Friday?") would increase the practical utility of the natural language interface for passengers with flexible itineraries. 
 
%% ============================================================
\section{Conclusion}
%% ============================================================
 
This paper presented FlightSense, an end-to-end MLOps platform for real-time flight delay prediction. Through progressive ablation across three model versions trained on 7.07 million BTS 2018 flight records, we demonstrated that delay propagation features derived from aircraft rotation chain reconstruction provide the dominant predictive gain (AUC 0.732 $\rightarrow$ 0.875, $+0.143$), while NOAA weather features contribute a meaningful incremental improvement (AUC 0.875 $\rightarrow$ 0.879). Feature importance analysis confirms the top four features are all rotation chain propagation variables, with \texttt{prev\_arr\_delay} dominating at gain 175.2, and snowfall ranking as the highest weather predictor at rank~6.
 
Baseline comparisons confirm that well-engineered features matter more than extensive hyperparameter tuning: a shallow untuned XGBoost with 27 engineered features (AUC 0.811) outperforms a fully tuned XGBoost with only 11 schedule features (AUC 0.732). The deployed system integrates live NOAA weather ingestion via AWS Lambda, real-time SageMaker inference, an interactive Streamlit dashboard, and an Amazon Bedrock Nova Micro conversational assistant that grounds delay risk estimates in live weather data through multiplicative probability compounding --- representing, to our knowledge, the first production deployment combining validated ML inference with an agentic conversational AI interface for flight delay prediction.
 
%% ============================================================

\end{document}